\pdfoutput=1
\documentclass[11pt]{article}

\usepackage{acl}

\usepackage{times}
\usepackage{latexsym}
\usepackage{amsmath}
\usepackage[linesnumbered,ruled,lined]{algorithm2e}
\usepackage{algpseudocode}
\usepackage{graphicx}
\usepackage{amssymb}
\usepackage{booktabs}
\usepackage{color}
\usepackage{makecell}
\usepackage{changepage}
\usepackage[T1]{fontenc}
\usepackage[utf8]{inputenc}

\usepackage{microtype}

\usepackage{xcolor}

\newcommand{\eat}[1]{{}}
\newcommand{\rere}[0]{\textsc{ReRe}}

\newcommand{\orgone}{\textsuperscript{$\diamondsuit$}}
\newcommand{\orgtwo}{\textsuperscript{$\heartsuit$}}
\newcommand{\orgthree}{\textsuperscript{$\spadesuit$}}
\newcommand{\orgfour}{\textsuperscript{$\clubsuit$}}
\renewcommand{\thefootnote}{\fnsymbol{footnote}}

\definecolor{yellow-arrow}{RGB}{255,205,40}
\definecolor{green-arrow}{RGB}{151,208,119}
\definecolor{blue-arrow}{RGB}{126,166,224}
\title{Adaptive Ordered Information Extraction with Deep Reinforcement Learning}
\author{Wenhao Huang\orgone, Jiaqing Liang\orgtwo \footnote[2]{} , Zhixu Li\orgone, Yanghua Xiao\orgone\orgthree\footnote[2]{} , Chuanjun Ji\orgfour\\
\orgone Shanghai Key Laboratory of Data Science, School of Computer Science, Fudan University \\
\orgtwo School of Data Science, Fudan University \\
\orgthree Fudan-Aishu Cognitive Intelligence Joint Research Center, \orgfour DataGrand Inc. \\
\texttt{whhuang21@m.fudan.edu.cn} \\ 
\texttt{\{liangjiaqing,zhixuli,shawyh\}@fudan.edu.cn} \\
\texttt{jichuanjun@datagrand.com}
}

\begin{document}
\maketitle
\footnotetext[2]{Corresponding authors.}

\renewcommand*{\thefootnote}{\arabic{footnote}}
\begin{abstract}
Information extraction (IE) has been studied extensively. The existing methods always follow a fixed extraction order for complex IE tasks with multiple elements to be extracted in one instance such as event extraction.
However, we conduct experiments on several complex IE datasets and observe that different extraction orders can significantly affect the extraction results for a great portion of instances, and the ratio of sentences that are sensitive to extraction orders increases dramatically with the complexity of the IE task.
Therefore, this paper proposes a novel adaptive ordered IE paradigm to find
the optimal element extraction order for different instances, so as to achieve the best extraction results. We also propose an reinforcement learning (RL) based framework to generate optimal extraction order for each instance dynamically. 
Additionally, we propose a co-training framework adapted to RL to mitigate the exposure bias during the extractor training phase.
Extensive experiments conducted on several public datasets demonstrate that our proposed method can beat previous methods and effectively improve the performance of various IE tasks, especially for complex ones. \footnote{Resources of this paper can be found at \url{https://github.com/EZ-hwh/AutoExtraction}}
\end{abstract}

\section{Introduction}
% 知识抽取是很重要的任务
Information Extraction (IE) has been studied extensively over the past few decades~\cite{grishman2019twenty}.
%
%Typical IE tasks include entity extraction~\cite{?}, relation extraction~\cite{?}, and event extraction~\cite{?} etc.
%
With the rapid development of pre-trained language models, simple IE tasks such as named entity recognition~\cite{nadeau2007survey, li2020survey} have been well solved. However, complex IE tasks with multiple elements to be extracted such as relation extraction~\cite{pawara2017relation} and event extraction~\cite{hogenboom2011overview} still need further exploration.

%The success of IE is essential for a variety of applications such as text parsing, semantic analysis, and knowledge graph construction.
%
%Due to the diversity of data sources, extraction targets, and extraction patterns, IE tasks could be very challenging.
%这里的extraction patterns指的是什么？是否属于技术路线，并不应该归于复杂原因？ 
% \wenhao{指的是不同的抽取模式，应该属于技术}
%}
% 我觉得第一段没必要继续介绍ie了 直接简单ie已经被大模型解决了  具有更多元素的复杂ie是现在需要考虑的
%
%Information extraction is a key step in knowledge graph construction. This task aims at identifying and structuring predefined schema information from unstructured texts. 
%
%its variety in different sources, different targets, different structures and different schema. With the success of pre-trained language models (PLMs) applying in natural language processing, current extraction framework achieve impressive performance while extracting role from a single sentence.

% 知识抽取的方法和框架上使用的是联合或者Fix顺序的抽取
% 这些方法都用到了固定的顺序
%
Traditional IE methods always follow a \emph{static} extraction manner, i.e. with a pre-defined fixed element order.
% : 分三类，分步，joint，generation
For instance, in relation extraction task, \cite{xie2021revisiting} recognizes the relation and then extract the subject and object independently. \cite{luan2019general} extracts all entity together and recognize the relation between them pair-wisely.
\cite{wang2020tplinker} formulates joint extraction as a token pair linking problem, which follows implicit predefined order.
\cite{lu2022unified} designs a unified structured extraction language to encode different IE structures, where the generation order defines the extraction order.
%follow a disordered extraction manner~\cite{}, i.e., extracting multiple elements independently.
%
These extraction-based IE methods assume that multiple elements of the same instance are independent of each other and thus will not affect each other's extraction results.

% 抽取顺序对句子的抽取结果存在较大的影响，with case
% We argue that ,....,
% 塞一个case进去
    % 黄丹仪给梅艳芳做过三首曲子，分别是《不快不吐》，《无名氏》，《爱的教育》，每一首都可以上梅艳芳代表作排行榜
    % 12月15日，赵真作词作曲，杜桦演唱的《旗袍美人》正式上线推广
\begin{figure}[t]
    \centering
    \includegraphics[width=\linewidth]{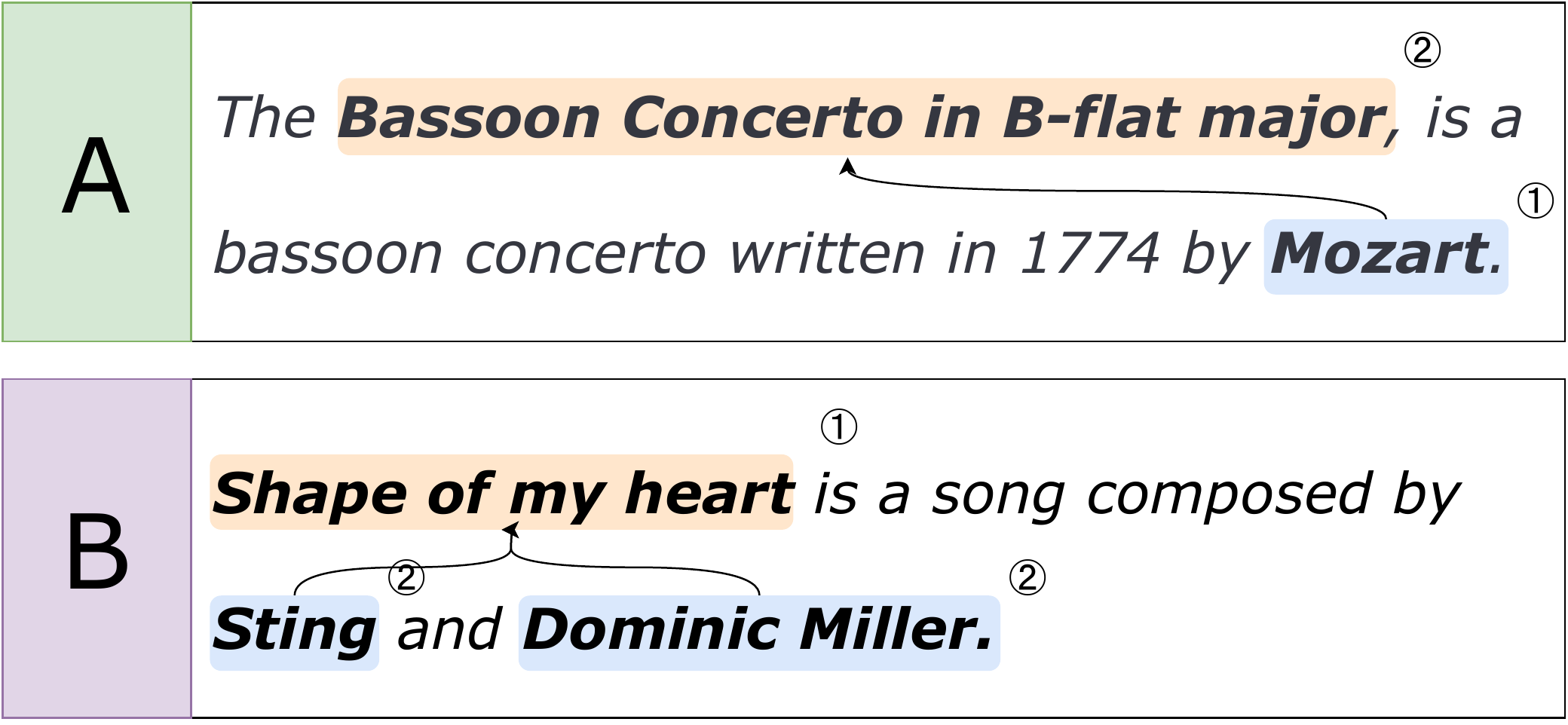}
    \caption{An example of complicated information extraction with different extraction order.}
    \label{fig:xmp4order}
\end{figure}

% 列指标，可以引RERE，说明顺序对抽取十分重要
\begin{table}[t]
    \centering
    \resizebox{1.05\columnwidth}{!}{
        \begin{tabular}{lccc}
        \toprule
            \textbf{Dataset} & \textbf{\#Ins.} & \textbf{\#Sens ins.} & \textbf{Ratio}  \\ 
        \midrule
            SKE21\cite{xie2021revisiting} & 1150 & 100 & 8.70\% \\
            NYT\cite{riedel2010modeling} & 5000 & 476 & 9.52\%  \\
            NYT10-HRL\cite{takanobu2019hierarchical} & 4006 & 525 & 13.11\%  \\ 
            DuIE\cite{li2019duie} & 15000 & 3618 & 24.12\%  \\ 
            DuEE\cite{li2020duee} & 1492 & 605 & 40.55\%  \\ 
            HacRED\cite{cheng2021hacred} & 1500 & 971 & 64.73\%  \\
        \bottomrule
        \end{tabular}
    }
    \caption{The statistic of sensitive instances on different datasets, where an instance is sensitive if different extraction orders produces different extraction results with the same model.}
    \label{tab:pilot}
\end{table}

% : 可以复用Figure，
% 例子一定要强，这样sensitive好解释
% : 用例子和统计指标来说明
However, we find that \emph{different extraction orders highly affect the extraction results.}
% FIXME: replace this example into a hard one
%\leeon{replace this example into a hard one ... For example, in the second case in Fig.~\ref{fig:xmp4order}, if we first extract the people, it will be difficult for the extractor because they play different role in different relation. On the contrary, however, the difficulty will be reduced.}
For example, in the A instance in Fig.~\ref{fig:xmp4order}, if we first extract the \textit{Bassoon Concerto in B-flat major}, it will be difficult for succeeding element extraction because the entity is long-tail. Instead, if we extract \textit{Mozart} first, it would be much easier to extract the concerto because \textit{Mozart} appears frequently in the corpus and he composed lots of concerto.
%\wenhao{In the event extraction, time is an essential element. Extracting time first can effectively help recognizing other elements of the events, such as places, people, etc. However, if people is extracted first, other element may be wrong or missing due to multiple different behaviors of people.}
%
According to our observation experiments conducted on several popular relation extraction and event extraction datasets as listed in Table~\ref{tab:pilot}, a significant proportion of sentences are sensitive to the extraction order, i.e., different extraction order produces different extraction results. 
% 不需要读数据。
%The ratio of sensitive sentences increases dramatically with the complexity of the task on different datasets, from 12.8\% on WebNLG and 24.12\% on DuIE, to as high as 64.73\% on HacRED and 74.80\% on DuEE\_fin.
The ratio of sensitive sentences increases dramatically with the complexity of the task on different datasets.

% 斜体 表示观点，和上一段相同
What's worse, \textit{the optimal extraction order may not be the same for different instances of the same relation or event type.}
%What's worse, for sentences having the same type of relation or event to be extracted, \textit{the optimal extraction order may not be the same.}
%
%\leeon{For instance, ... modify this example after you change fig.1.}
% FIXME: modify this example after you change fig.1.
For example, for the \textit{composer} relation, we should extract the composer \textit{Mozart} first in A instance, but extract the song \textit{Shape of my heart} first in B instance in Fig~\ref{fig:xmp4order} because of its frequent appearance in corpus.
Based on the observations above, the \emph{static} extraction paradigm following a pre-defined extraction order is insufficient to achieve reliable extraction.
This motivates us to propose a \emph{dynamic} extraction paradigm, which aims to assign an optimal element extraction order for each instance adaptively. %However, few extraction frameworks focus on the decision of extraction order.
%A \emph{dynamic} extraction framework is better. However, few extraction frameworks are able to adaptively use the right order of extraction for the right sentence.
%
%\wenhao{Based on the observations above, this paper proposes a novel adaptive ordered IE paradigm to find the optimal element extraction order for different instances, so as to achieve the best extraction results.}
%
%The challenges lie on how to find the optimal extraction order for different instances.
%
%多几句话来讲述这个难点，最好分2-3点，然后在下一段可以呼应起来。
%
It is nontrivial to dynamically find optimal extraction order for every instance in a dataset.
%
%\leeon{First, each sentence contains different elements to be extracted and they should be distinguished....modify}
%
On one hand, the optimal extraction order of an instance depends on not only its schema (relation or event type), but also the context of the sentence.
On the other hand, multiple rounds of decisions are required to generate the optimal extraction order, where the decision of each step depends on not only the schema and sentence context, but also the extraction results of previous steps.
% 描述一下怎么样的场景适合使用RL
%Since extraction order is determined dynamically, multiple rounds of decisions are needed, which can be solved by reinforcement learning.
%\wenhao{So we use RL to figure out this problem.}
%in addition to depending on the sentence itself, the extraction order should also rely on the extracted elements, which can help the following extraction. \wenhao{Easily extractable arguments should be extracted first, and the illegible ones should refer to the previously extracted results for assistance.}
%
%Third, since the extraction order is determined dynamically, multiple rounds of decisions are needed. \wenhao{The extraction order not only depends on the sentence and the corresponding Schema, but also depends on the previous extraction results.}

% FIXME: 具体方法:可以略微简化，
%To address the above obstacles
In this paper, we propose to adopt value-based reinforcement learning~\cite{mnih2015human} in determining the optimal extraction order for elements of an instance.
%
%Particularly, for each instance, the value of each unextracted elements is evaluated according to the sentence and all the previously extracted arguments, and then the role with the highest value is extracted in the next round.
%
Particularly, in deciding the next extraction element for an instance, every of its unextracted elements will be evaluated with a potential benefit score, which is calculated with a BERT-based model.
Then, the one with the highest potential benefit score will be selected as the next extraction object.
In addition, to mitigate the exposure bias that emerges during the RL-based extractor training phase, a co-training framework is adopted which can simulate the inference environment of the extraction order decision agent to help enhance its performance.
It is worth mentioning that our method focuses on generating the optimal extraction order, which is model agnostic and can be applied to various extraction paradigms.

Our main contributions are summarized below:
\begin{itemize}
    \item First, we propose the extraction order assignment problem in complicated IE task, which can effectively affect the extractor and the extraction result.
    \item Second, we propose an RL based framework that dynamically generates an optimal extraction order for each sentence, which is model agnostic and can effectively guide the model towards better extraction performance.
    \item Third, we adopt a co-training framework for RL to alleviate the exposure bias from the extractor, which can simulate the inference environment of the extraction order decision agent to help enhance its performance.
    \item Fourth, the experiments conducted on several public datasets show that our method outperforms the state-of-the-art extraction models.
\end{itemize}

\section{Related Work}
% 1. 当前针对复杂知识抽取的框架分类及缺点
% 关系抽取的相关工作
% : 更改一种表述

Pipeline Information Extraction (IE) methods split the extraction process into several sub-tasks and optimize each of them. They rely on the task definition so the framework varies for different IE tasks. For relation extraction task, \cite{wei2020novel,xie2021revisiting,li2021tdeer} gradually extract subject, relation and object from the sentence in different order. For event extraction task, \cite{yang2018dcfee, sheng2021casee, yang2019exploring} first recognize the event type and trigger and then extract the arguments with sequence tagging model or machine comprehension model.

Joint IE methods combine two or more extraction processes into one stage. Graph-based methods are the mainstream joint IE framework. They recognize the entity or text span and build a graph with co-reference, relation~\cite{wadden2019entity,luan2019general}, entity similarity~\cite{xu2021document} or sentence (co-occurrence)~\cite{zhu2021efficient}. Through information propagation on the graph, they better encode the sentence and document and then decode the edge to build the final sub-graph. Generation-based IE methods are another paradigm for joint extraction, \cite{cabot2021rebel,ye2021contrastive} for relation extraction task, \cite{zheng2019doc2edag,hsu2022degree,du2021template} for event extraction, \cite{lu2022unified} for unified information extraction, they all serialize the structured extraction results into sentence or pre-defined template in a fixed order. Apart from works above, \cite{wang2020tplinker, shang2022onerel} propose one-stage joint extraction framework that decode the subject and object simultaneously.

% 2. 它们为何不能处理不同顺序的抽取问题

% 强化学习
Recently, reinforcement learning (RL) has been applied to IE task. Information extraction was augmented by using RL to acquire and incorporate external evidence in \cite{narasimhan2016improving}. \cite{feng2018reinforcement,wang2020rh} both train a RL agent for instance selecting to denoise training data obtained via distant supervision for relation extraction. \cite{takanobu2019hierarchical} utilizes a hierarchical reinforcement learning framework to improve the connection between entity mentions and relation types. \cite{zeng2019learning} first considers extraction order of relational facts in a sentence and then trains a sequence-to-sequence model with RL.

% 字体换下  压完好丑
\begin{figure}[tb]
    \centering
    \includegraphics[width=\linewidth]{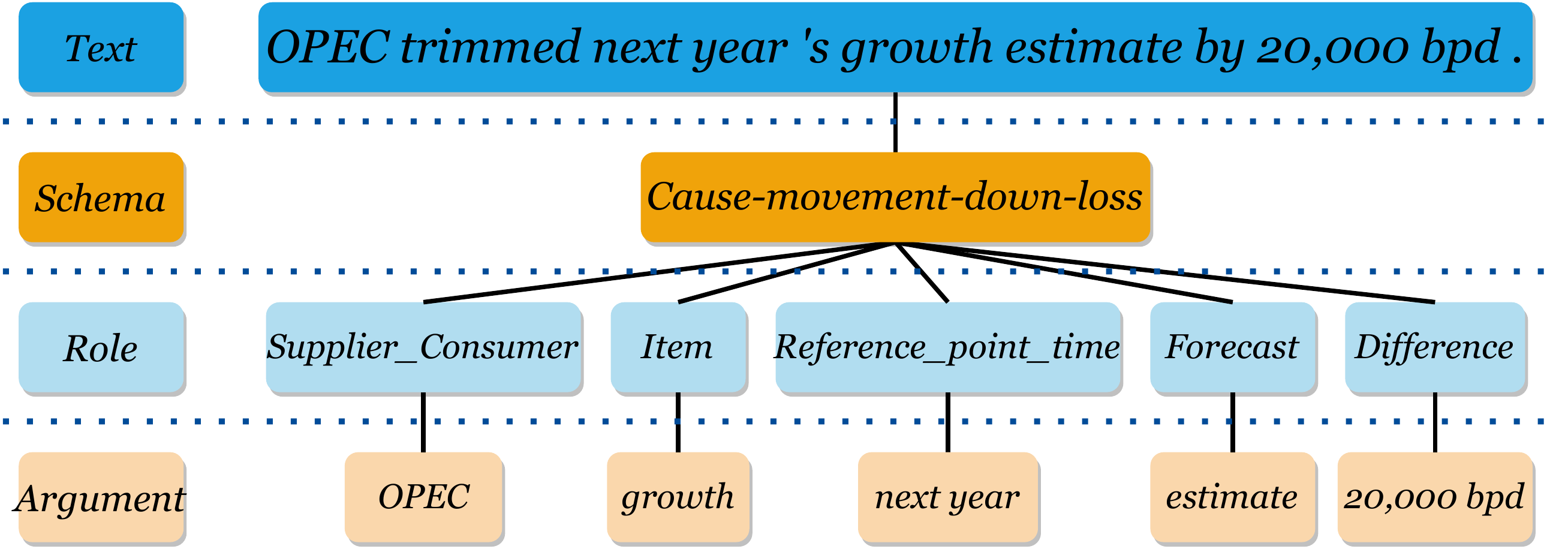}
    \caption{An example of Complicated Information Extraction}
    \label{fig:xmp}
\end{figure}

% 学order是天然的RL建模
% Overview 太细了  细节移到sec 4

\section{Overview}
\subsection{Task Definition}
% 复杂抽取任务的形式定义
Fig. \ref{fig:xmp} gives an example of complicated information extraction process, which first recognizes the schema and then extracts the argument $\tt arg_i$ for the corresponding role $\tt role_i$. Generally speaking, the task can be split into two sub-tasks: relation (event) detection and entity (argument) extraction.
And we formulate the second task as multi argument extraction task. Given an instance $s$, the relation/event type $rel$ and the corresponding pre-defined schema $\tt <rel, role_1, role_2, ...,role_n>$, our goal is to find all the arguments in $s$ and fill them in their corresponding roles in schema. 

% 为什么用RL
% 本质上是序列问题

% 多轮要素的抽取建模
% 提及我们的顺序抽取是Model Agnostic的。
% 统一一下表述common,simple,difficult,frequent之类的表述需要解释
\subsection{Solution Framework}

In this work, we model the complicated information extraction as a multi-step argument extraction task. It this setting, only a role in the schema will be extracted from instance once. With the help of extractor that can extract the arguments given the additional information and role name, we extract all arguments and fill the roles step by step. %What

Though there are various roles in complicated IE task, the difficulty of extracting them are completely different. For example, the role \textit{Reference\_point\_time} in Fig~\ref{fig:xmp} indicates the time in the context and it can be extracted without further information. Other roles like \textit{Supplier\_Consumer}, however, can not be identified with a single role name, so they should be scheduled for extraction later.
%So The argument that we extract from the common role would be a supplementary information helping extractor.} %Why
We hope that the extractor can first extract the simplest role, then the next simplest one and etc. By incrementally adding the previously extracted information, the extractor can keep a good performance on extracting the difficult ones. %How

% 为什么建模成多轮
%为什么使用RL
To achieve the goal, we need to arrange a reasonable extraction order for the extractor. However, it is hard to specify the whole extraction order once because it depends on not only the schema and context, but also the previous extracted arguments. So we regard the extraction order decision as a \textit{Markov decision process}, where we can dynamically decide the extraction order in multi-round. Clearly, reinforcement learning is a natural way to handle this modeling. We adopt the double deep Q-network (DQN) with prioritized replay buffer as RL agent.
%which we model as \textbf{extraction order decision task}. 
%Since there are multiple schema in an extraction task, the difficulty of extracting a role varies in different instances. 
%Confirming a certain extraction order for all sentence or schema is not satisfactory in many cased. %Why
%We solve the problem using a reinforcement learning framework that is illustrated in the next section.%How
%\textit{
%Clearly, reinforcement learning is a natural way to determine the dynamic extraction order.
%}

\section{Framework}
\begin{figure*}[!htb]
    \centering
    \includegraphics[width=\textwidth]{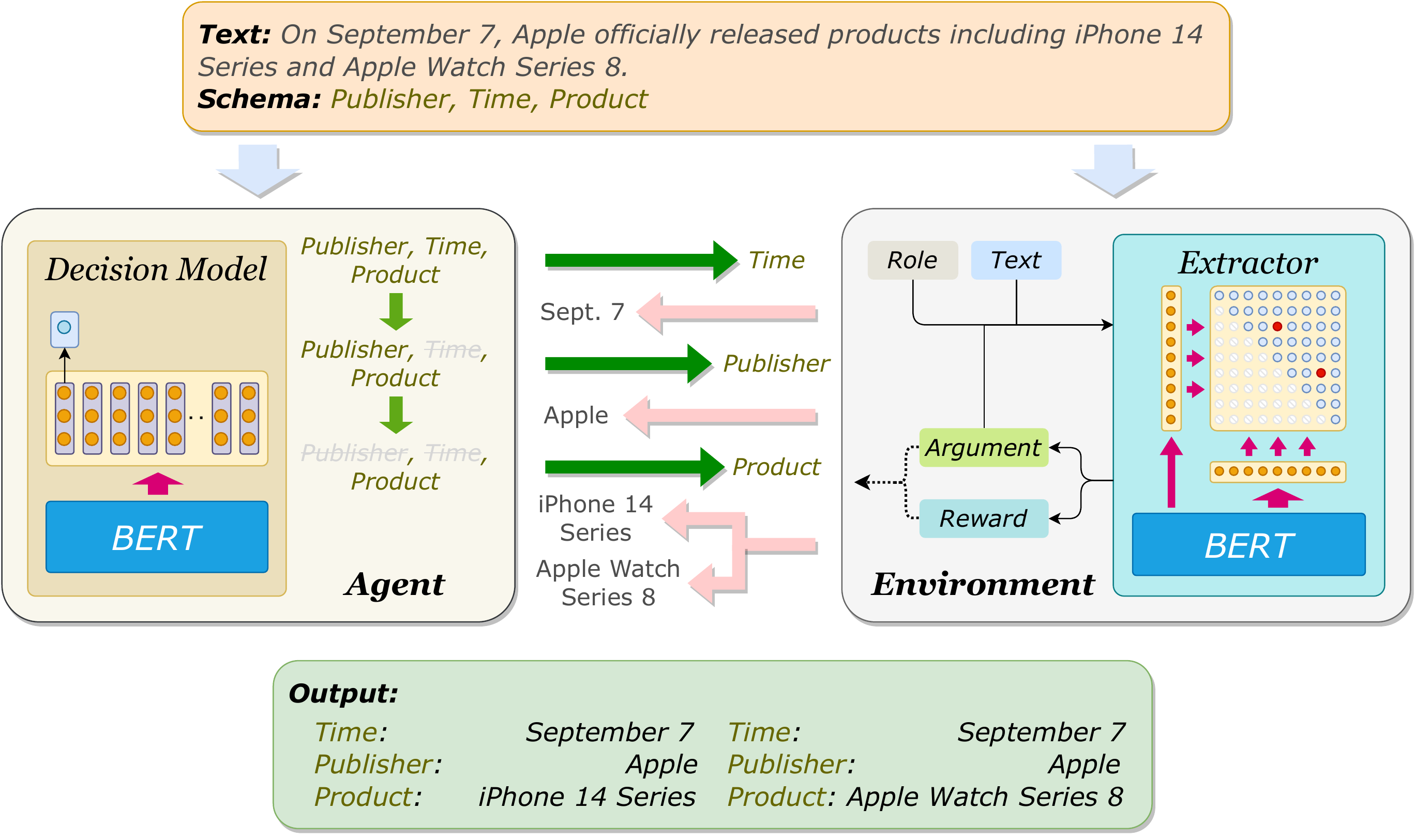}
    \caption{Reinforcement learning framework of adaptive ordered decision for IE.}
    \label{fig:framework}
\end{figure*}

%We first introduce our basic model and then introduce the environment setting where RL agent is applied on. Our decision model is 
\subsection{Extractor}
\label{sec:extractor}
% 要说明为什么使用这个抽取器，同时说明抽取器本身已经能有很好的抽取效果
% 针对不同的条件输入，它能有不同的输出, 同时能有实体区间的打分
% 
To handle the extraction tasks with different extraction order, we have to use a powerful extractor. GlobalPointer, proposed by \cite{su2022global}, is an ideal choice as it can identify both nested and non-nested entities, and even output the scores of the entities.

We first construct the sequence consisting of the extracted elements, role name and the sentence. For an input sequence with $N$ tokens, the BERT based encoder outputs a vector sequence $[\textbf{h}_1, \textbf{h}_2, ..., \textbf{h}_N]$. Following the computation of attention matrix, we use a one-head self-attention to compute the matrix as the output of the decoder. More specifically, we first convert the vectors $\textbf{h}_i$ to vectors $\textbf{q}_i$ and $\textbf{k}_i$ with two linear layers. 
\begin{equation}
    \begin{aligned}
        \textbf{q}_i &= \textbf{W}^q\textbf{h}_i + \textbf{b}^q, \\
        \textbf{k}_i &= \textbf{W}^k\textbf{h}_i + \textbf{b}^k,
    \end{aligned}
\end{equation}
where $\textbf{W}$ and $\textbf{b}$ are parameters of the linear layers. Then we compute the scores of each spans with the with relative position embeddings (RoPE) proposed by Reformer \cite{kitaev2020reformer}. The transformation matrix $\textbf{R}_i$, which compose of sine and cosine function, satisfy the property that $\textbf{R}_i^\top \textbf{R}_j=\textbf{R}_{j-i}$. By introducing relative positions embeddings, GlobalPointer is more sensitive to the length and span of entities, so it can better distinguish real entities.

\begin{equation}
    \begin{aligned}
        s_\alpha (i,j) &= \left(\textbf{R}_i\textbf{q}_{i}\right)^\top \left(\textbf{R}_j\textbf{k}_j\right) \\
        & = \textbf{q}_i^\top \textbf{R}_i^\top \textbf{R}_j \textbf{k}_i \\
        & = \textbf{q}_i^\top \textbf{R}_{j-i} \textbf{k}^j
    \end{aligned}
\end{equation}

To better solve the label imbalance problem, we use the following loss to train our extraction model.
\begin{equation}
    \begin{aligned}
        L = &\log\left(1+\sum_{(i,j)\in P_\alpha} e^{-s_{\alpha} (i,j)}\right) + \\
        & \log\left(1+\sum_{(i,j)\in Q_\alpha} e^{s_{\alpha} (i,j)}\right)
    \end{aligned}
\end{equation}
where $P_\alpha$ is the head-tail set of all spans of query $\alpha$, and $Q_\alpha$ is the head-tail set of the rest valid spans in text. In the decoding phase, all spans $t_{[i:j]}$ for which $s_\alpha(i,j) > 0$ are considered as entities that match the conditions.

For better fitting into the setting of extraction tasks that extract entities under the conditions of the schema and role name, we enumerate all the extraction orders and match the corresponding conditions with the extraction results to construct a training instance.

\subsection{MDP for extraction order}
%如何将抽取顺序决策问题建模成MDP问题
We regard the multi-role extraction order decision process as a \textit{Markov decision process} (MDP). Fig.~\ref{fig:framework} shows the whole extraction process of an instance. In each step, the agent takes the instances and extracted arguments as input, and chooses a role unselected before as the action. The environment would take the selected role, and construct the input sequence for extractor. After collecting the extraction results to fill the role and extraction scores to assign the reward, the environment would transit to new state(s). After selecting all roles to be extracted in multiple rounds, we exchange the whole extraction history into structural output.

% sec4 framework 放这里
% 后面全丢sec 4
\paragraph{State}
We use $s_t$ to denote the state of sentence $x$ in extracting time step $t$. The state $s_t$ consists of the extraction schema $\mathcal{S}$, the already extracted arguments $\hat{y}^{<t}$ and the sentence $x$.
\begin{equation}
    s_t = (\mathcal{S}, \hat{y}^{<t}, x)
\end{equation}
The state describes the extracted element in the past step. In each step, the environment would take a role selected by the agent and extracts the corresponding arguments in the sentence with the help of extractor described in Section \ref{sec:extractor}.

\paragraph{Action}
The action of the RL model is the next role to extract in an instance. Unlike the traditional RL environment, the action space in our model is continuously reduced at every time step. We restrict the initial action space $\mathcal{A}_0$ to the set of roles in the schema $\mathcal{S}$. After selecting role $a_0$ in time step $0$, the extractor in environment will extract the argument and its confident score in $s$ with the help of extractor. The action $a_0$ will be removed from the $\mathcal{A}_0$ and derive the next action space $\mathcal{A}_1$. The derivation of action space can be formalized as below.
\begin{equation}
    \mathcal{A}_t = 
    \begin{cases}
            \mathcal{S}, & t = 0\\
            \mathcal{A}_{t-1} - \{a_{t-1}\}, & 0<t<|\mathcal{S}| 
    \end{cases}
\end{equation}

%\begin{figure}[tb]
%    \centering
%    \includegraphics[width=\linewidth]{figure/MDP Process.pdf}
%    \caption{MDP process for extraction order decision task.}
%    \label{fig:mdp}
%\end{figure}

% Intermediate Reward + terminated reward.
\paragraph{Reward}
The reward is a vital important component of RL training, which is an incentive mechanism that encourages agent to better plan for the episode. For our task definition, there is a simple reward assignment. We can extract all the arguments in the sentences, and then assign a reward in terminated state to indicate whether the extracted tuple matches the golden label. But there is a transparent issue that it will majorly depends on the extractor we use. If the extractor is too strong, the results extracted following any extraction order are correct. If the extractor is too weak, the results extracted following any extraction order are incorrect. In the cases described above, different extraction order can not affect the final reward. Therefore, to better distinguish the impact of different extraction orders on the extractor, we define the reward as the score of the extraction results by the extractor. Though the extracted results recognized by the extractor for a single step is the same, the score is different given the different condition according to the Section \ref{sec:extractor}. We regard the score as the difficulty of extracting the argument from the sentence. An extracted argument with high score indicates that it is easy to be extracted. The reward of our RL environment can be described as below.

% Extractor 的独立性 说明  怎么产生分数
\begin{equation}
\mathcal{R}(s,a) = Extractor_{score}(a|s)
\end{equation}

where $s$ stands for the state and $a$ stands for the role that will be extracted chosen by the agent.

\begin{algorithm}[ht]
    \caption{The full details of our training phase for the Double DQN agent with $\epsilon-$greedy exploration}
        \DontPrintSemicolon
        \SetKwInOut{Input}{Input}
        \SetKwInOut{Output}{Output}
        \SetKwFunction{FRandomMain}{Random}
        \SetKwFunction{FRandomS}{Random-Sample}
        \Input{$\mathcal{D}$-empty replay buffer; $\theta$-initial network parameters; $\theta^-$-copy of $\theta$}
        
        \Input{$N_b$-training batch size; $N^-$-target network replacement frequency}

        \For{epoch $= 1,...,E$}{
            Sample instances $s,\mathcal{S}$ from the dataset.
            
            $N_{step}$ = \#number of roles in the $\mathcal{S}$
            
            \For{t $= 1,...,N_{step}$}{
                $p \gets$ \FRandomMain{$0,1$}
                
                \eIf{p < $1 - \epsilon$}{
                    $a_t \gets \arg\max_aQ(s_t,a;\theta)$
                }{
                    $a_t \gets$ \FRandomS{$\mathcal{A}_t$}
                    
                }
                
                $s_{t+1}, r_t \gets Transition(s_t,a_t)$
                
                Store transition $(s_t,a_t,r_t,s_{t+1})$ in $\mathcal{D}$
                
                Sample random mini batch of $N_b$ transitions $(s_t,a_t,r_t,s_{t+1})$ from $\mathcal{D}$
                
                \eIf{$s_{t+1}=done$}
                {
                    $y_t=r_t$
                }{
                    $y_t=r_t+\gamma\max_{a'}Q(s_{t+1},a';\theta^-)$
                }
                Update parameter $\theta$ on the loss $\mathcal{L}(\theta)=(y_t-Q(s_t,a_t;\theta))^2$

                Replace target parameters $\theta^- \gets \theta$ every $N^-$ steps
            }
        }
    \label{alg:training framework}
\end{algorithm}

\subsection{Double DQN}
For traditional DQN\cite{mnih2013playing}, the learning object follows the Bellman equation as below.
\begin{equation}
    Q(s,a) = \mathcal{R}(s,a) + \gamma\cdot \max_{a' \in \mathcal{A}}Q(s',a')
\end{equation}

In our task setting, the agent would choose a unextracted role and the environment would return the extracted argument with the corresponding score. Since the extractor will extract all the corresponding entities that meet the conditions at once, it is possible to extract zero to multiple answers at one time, and each extracted result will form a separated state. Due to the splitting of the state, we need to make corresponding adjustments to the original Learning object. Inspired by \cite{tavakoli2018action}, we introduce a new learning object adapted to the branching reinforcement learning environment by replacing the $Q$ values of the next state with the average value of the next state's list.
% 插入一个分支强化学习的示意图，用于解释状态的变化
\begin{equation}
    Q(s,a) = \mathcal{R}(s,a) + \gamma\frac1{|S_{(s,a)}|}\sum_{s'\in S_{(s,a)}}{\max_{a' \in \mathcal{A}}Q(s',a')}
    %Q(s,a) = \mathcal{R}(s,a) + \gamma \max_{a' \in \mathcal{A}}{\frac1{N_{(s,a)}}\sum_{s'\in\mathcal{S}}Q(s',a')}
\end{equation} % 符号解释需要完善
where $S_{(s,a)}$ is the set of the following states derived from the state $s$ with action $a$, and $\gamma$ is the discount factor.

%\subsection{Reinforcement Learning Process}
% Double DQN
% Replay buffer

%\wenhao{There exist numerous ways by which the distributed TD errors across the branches can be aggregated to specify a loss. A simple approach is to define the loss to be the expected value of a function of the averaged TD errors across the branches. However, due to the signs of such errors, their summation is subject to canceling out which, in effect, generally reduces the magnitude of the loss. To overcome this, the loss can be specified as the expected value of a function of the averaged absolute TD errors across the branches. In practice, we found that defining the loss to be the expected value of the mean squared TD error across the branches mildly enhances the performance:}

To avoid suffering from the overestimation of the action values, we adopt the Double DQN (DDQN) algorithm \cite{van2016deep} that uses the current Q-network to select the next greedy action, but evaluates it using the target. At the same time, to enable more efficient learning from the experience transitions, we adopt the prioritized experience replay buffer \cite{schaul2015prioritized} to replay important experience transitions.

We define the loss to be the expected value of the mean squared TD error.
\begin{equation}
    \mathcal{L} = \mathbb{E}_{(s,a,r,s')\sim\mathcal{D}}\left[y - Q(s,a)\right]^2
\end{equation}
where $\mathcal{D}$ denotes the prioritized experience replay buffer and $y$ denotes the estimated valued by the target network.

To evaluate the value of (state, action) pair, we use Transformer based model as encoder that can take the state and action as input and encode the pair of every (state, action) pair. Specifically, we use BERT\cite{devlin2018bert} for English and RoBERTa\cite{liu2019roberta} for Chinese. Formally, for an input state $\textbf{s}_t = [t_1, t_2, ..., t_N]$ and action $\textbf{a}_t = [a_1, ..., a_M]$, where the action is the candidate extraction role name, we form the sequence $x = \left[\text{\tt[CLS]}, \textbf{a}_t, \text{\tt[SEP]}, \textbf{s}_t, \text{\tt[SEP]}\right]$. The BERT encoder converts these tokens into hidden vector $[\textbf{h}_1, \textbf{h}_2, ..., \textbf{h}_{M+N}]$, where $\textbf{h}_i$ is a $d$-dimension vector and $d=768$ in the Transformer based structure.

% 如何使用Q(S,A)进行值的预测，
To evaluate the $Q(s,a)$ value for the corresponding state and action, we take $\textbf{h}_0$, which is the encoded vector of the first token $\text{\tt[CLS]}$ as the representation of the state-action pair. The final output of the value evaluation module $\hat{\textbf{y}}$ is define in Eq.
\begin{equation}
    \hat{\textbf{y}} = \textbf{W}\textbf{h}_0+\textbf{b}
\end{equation}
where $\textbf{W}$ and $\textbf{b}$ are trainable model parameters, representing weights and bias of the linear transformation.

\begin{figure}[tb]
    \centering
    \includegraphics[width=\linewidth]{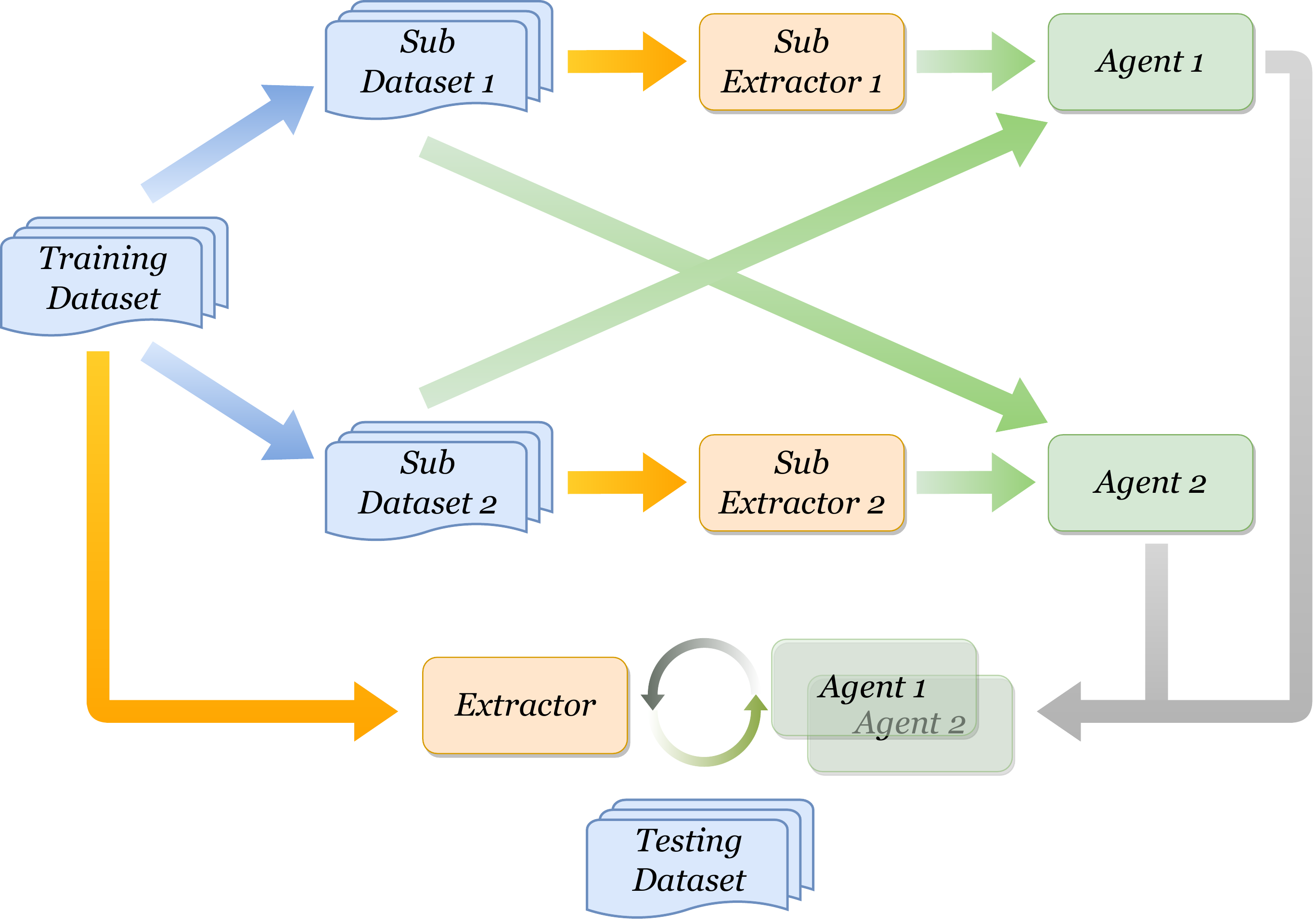}
    \caption{Framework of co-training. \textcolor{blue-arrow}{Blue} arrow represents the training data division process. \textcolor{yellow-arrow}{Yellow} arrow represents the extractor training process. \textcolor{green-arrow}{Green} arrow represents the RL agent training process.}
    \label{fig:cct}
\end{figure}

\subsection{Co-training framework}
%Why: training和testing的环境存在不同。
Beside the extraction order our agent decide would affect the final extraction order, the extractor in the environment also matters. The argument extracted in one step not only affect the extracted tuple, but also affect the decision of the agent. However, there is a big different between the training phase and inference phase. In the training phase, the agent explores in the environment that the extractor has a good performance, which will extract the argument with high score. In the inference phase, however, the capacity of the extractor would be reduced because of the migration of the dataset from training to testing. To confirm that the agent would works in inference phase, we need to ensure similarity between the training environment and the inference environment.

%How  1. 将数据集随机切分成两个大块，然后将分别训练两个子抽取器；2. 将两个子抽取器交叉分别训练子Agent
We proposed a \textbf{Co-training} framework as shown in Fig \ref{fig:cct} to simulate the environment in the testing phase. We first split the training set into two piece, which are used to train two sub extractor. Then we crossover the sub dataset and sub extractor and build two training environments. In each environment, the extractor is trained on the other piece of sub training set and extracts arguments from the sentence that it never meets. We train two agent with two environment separately. By introducing the co-training framework, we confirm the same setting on the training and inference environment. In the inference phase, we can build the environment with the test set and the extractor trained on the whole training set. Through combining the decision that two agent make, the argument in the sentences can be extracted step by step.

% : 主实验结果是否需要融入前一步的分类结果 -> 需要
\begin{table*}[!htb]
	\centering
	\label{tab:apmainen}
	\resizebox{2.1\columnwidth}{!}{
		\begin{tabular}{p{5cm}rrrrrrrrrrrr}
			\toprule
			& \multicolumn{3}{c}{NYT}  & \multicolumn{3}{c}{NYT10-HRL}  & \multicolumn{3}{c}{HacRED} & \multicolumn{3}{c}{SKE21}      \\
			\cmidrule{2-13}             & \multicolumn{1}{c}{Prec.} & \multicolumn{1}{c}{Reca.} & \multicolumn{1}{c}{F1} & \multicolumn{1}{c}{Prec.} & \multicolumn{1}{c}{Reca.} & \multicolumn{1}{c}{F1} & \multicolumn{1}{c}{Prec.} & \multicolumn{1}{c}{Reca.} & \multicolumn{1}{c}{F1} & \multicolumn{1}{c}{Prec.} & \multicolumn{1}{c}{Reca.} & \multicolumn{1}{c}{F1} \\
			\midrule
			NovelTagging \# \cite{zheng2017joint} & 62.4 & 31.7 & 42.0 & 59.3     & 38.1     & 46.4 & 30.51 & 2.91 & 5.31 & \multicolumn{1}{c}{-} & \multicolumn{1}{c}{-} & \multicolumn{1}{c}{-} \\
			CoType \# \cite{ren2017cotype} & 42.3 & 51.1 & 46.3 & 48.6     & 38.6     & 43.0       & \multicolumn{1}{c}{-} & \multicolumn{1}{c}{-} & \multicolumn{1}{c}{-} & \multicolumn{1}{c}{-} & \multicolumn{1}{c}{-} & \multicolumn{1}{c}{-}\\
			CopyR \# \cite{zeng2018extracting} & 61.0 & 56.6 & 58.7 & 56.9     & 45.2 & 50.4 & 13.11 & 9.64 & 11.12 &\multicolumn{1}{c}{-} & \multicolumn{1}{c}{-} & \multicolumn{1}{c}{-}  \\
			HRL \# \cite{takanobu2019hierarchical}  & \multicolumn{1}{c}{-} & \multicolumn{1}{c}{-} & \multicolumn{1}{c}{-} & 71.4     & 58.6     & 64.4  & \multicolumn{1}{c}{-} & \multicolumn{1}{c}{-} & \multicolumn{1}{c}{-} & \multicolumn{1}{c}{-} & \multicolumn{1}{c}{-} & \multicolumn{1}{c}{-} \\
                CasRel \# \cite{wei2020novel} & 89.7 & 89.5 & 89.6 & 77.7 & 68.8 & 73.0 & \underline{55.24} & \underline{43.78} & \underline{48.85} & \multicolumn{1}{c}{-} & \multicolumn{1}{c}{-} & \multicolumn{1}{c}{-}  \\
			TPLinker \# \cite{wang2020tplinker}  & \underline{91.3} & \underline{92.5} & \underline{91.9} & \underline{81.19} & 65.41 & 72.45 & \multicolumn{1}{c}{-} & \multicolumn{1}{c}{-} & \multicolumn{1}{c}{-} & \multicolumn{1}{c}{-} & \multicolumn{1}{c}{-} & \multicolumn{1}{c}{-} \\
			\rere \# \cite{xie2021revisiting} & \multicolumn{1}{c}{-} & \multicolumn{1}{c}{-} & \multicolumn{1}{c}{-} & 75.45 & \underline{72.50} & \underline{73.95} & \multicolumn{1}{c}{-} & \multicolumn{1}{c}{-} & \multicolumn{1}{c}{-} & \multicolumn{1}{c}{-} & \multicolumn{1}{c}{-} & \multicolumn{1}{c}{-}\\
			%\midrule
			%CasRel \cite{wei2020novel} & - & - & - & - & - & - & - & - & -    \\
			% \rere - LSTM (exact) & 53.99    & 39.99    & 45.95    & 52.59    & 32.97    & 40.53    & 57.33    & 34.19    & 42.83    & 69.9     & 53.65    & 60.71 \\
			%\rere  \cite{xie2021revisiting} & - & - & - & - & - & - & - & - & -  \\
			\midrule
			\midrule
   
			CasRel\cite{wei2020novel} * & 87.77 & 90.79 & 89.25 & 76.59 & 68.90 & 72.54 & 57.19 & 44.99 & 50.36 & 87.21 & 75.23 & 80.78\\
   
			TPLinker\cite{wang2020tplinker} * & 88.61 & 92.29 & 90.41 & \textbf{80.37} & 65.11  & 71.94 & \textbf{58.96} & 55.78 & 57.33 & 83.86 & 84.77 & 84.32  \\
			
			\rere\cite{xie2021revisiting} * & 85.68 & 92.45 & 88.93 & 74.43 & 68.46 & 71.32 & 46.42 & 61.37 & 52.86  & 85.65 & 86.37 & 86.01 \\
   
            Adaptive Order & \textbf{88.92} & \textbf{92.83} & \textbf{90.84} & 77.21 & \textbf{69.81} & \textbf{73.32} & 58.36 & \textbf{72.43} & \textbf{64.64} & \textbf{87.99} & \textbf{86.87} & \textbf{87.42}\\
			%& & & &  &  &  & &  &  \\
			\bottomrule
		\end{tabular}%
  }
  \caption{The main evaluation results of different models on NYT, NYT10-HRL, HacRED and SKE21. The results with only one decimal are quoted from~\protect\cite{wei2020novel}. 
    The methods with * are based on our re-implementation. 
    The methods with \# denote that the metrics are partially matched.
     Best exact (partial) match F1 scores are marked \textbf{bold} (\underline{underlined}).}
    \label{tab:main}
\end{table*}%

\begin{table*}[!htb]
    \centering
    \resizebox{2.1\columnwidth}{!}{
    \begin{tabular}{lrrrrrrrrrrrrrrr}
    \toprule
         & \multicolumn{3}{c}{NYT} & \multicolumn{3}{c}{NYT10-HRL} & \multicolumn{3}{c}{HacRED} & \multicolumn{3}{c}{DuIE} & \multicolumn{3}{c}{DuEE} \\
         \cmidrule{2-16}       & \multicolumn{1}{c}{Prec.} & \multicolumn{1}{c}{Reca.} & \multicolumn{1}{c}{F1} & \multicolumn{1}{c}{Prec.} & \multicolumn{1}{c}{Reca.} & \multicolumn{1}{c}{F1} & \multicolumn{1}{c}{Prec.} & \multicolumn{1}{c}{Reca.} & \multicolumn{1}{c}{F1} & \multicolumn{1}{c}{Prec.} & \multicolumn{1}{c}{Reca.} & \multicolumn{1}{c}{F1} & \multicolumn{1}{c}{Prec.} & \multicolumn{1}{c}{Reca.} & \multicolumn{1}{c}{F1} \\
        
    \midrule
     Sequence order & 92.83 & \textbf{96.19} & 94.48 & \textbf{84.66} & 85.42 & 85.04 & \textbf{58.18} & 79.76 & 67.28 & 52.14 & 62.07 & 56.68 & 72.70 & 70.47 & 71.57\\
     Random order & 93.05 & 95.97 & 94.49 & 84.42 & 85.08 & 84.75 & 57.82 & 80.02 & 67.13 & 52.53 & 62.11 & 56.92 & 72.86 & 71.10 & 71.97\\
     Adaptive order & \textbf{93.15} & 96.14 & \textbf{94.62} & 84.59 & \textbf{85.66} & \textbf{85.13} & 57.88 & \textbf{81.92} & \textbf{67.62} & \textbf{53.33} & \textbf{62.87} & \textbf{57.71} & \textbf{73.86} & \textbf{72.14} & \textbf{72.99} \\
    \bottomrule
    \end{tabular}} 
     \caption{Extraction Result on different dataset with different extraction order decision. }
     \label{tab:order}
\end{table*}

%\begin{table*}[!htb]
%    \centering
%    \resizebox{2.1\columnwidth}{!}{
%    \begin{tabular}{p{3.2cm}rrrrrrrrrrrrrrr}
%    \toprule
%         & \multicolumn{3}{c}{SKE} & \multicolumn{3}{c}{NYT10-HRL} & \multicolumn{3}{c}%{HacRED} & \multicolumn{3}{c}{DuIE} & \multicolumn{3}{c}{DuEE} \\
%         \cmidrule{2-16}       & \multicolumn{1}{c}{Prec.} & \multicolumn{1}{c}{Reca.} & \multicolumn{1}{c}{F1} & \multicolumn{1}{c}{Prec.} & \multicolumn{1}{c}{Reca.} & \multicolumn{1}{c}{F1} & \multicolumn{1}{c}{Prec.} & \multicolumn{1}{c}{Reca.} & \multicolumn{1}{c}{F1} & \multicolumn{1}{c}{Prec.} & \multicolumn{1}{c}{Reca.} & \multicolumn{1}{c}{F1} & \multicolumn{1}{c}{Prec.} & \multicolumn{1}{c}{Reca.} & \multicolumn{1}{c}{F1} \\
        
%    \midrule
%     Sequence Extraction & 94.74 & 81.08 & 87.38 & - & - & - & 66.62 & 78.67 & 72.15 & 44.88 & 48.67 & 46.70 & 72.70 & 70.47 & 71.57\\
%     Random Extraction & 92.75 & 80.63 & 86.27 & - & - & - & 65.94 & 79.41 & 72.05 & 44.25 & 47.91 & 46.01 & 72.86 & 71.10 & 71.97\\
%     Auto Extraction & 93.37 & 82.43 & 87.56 & - & - & - & 66.31 & 80.56 & 72.74 & 46.60 & 50.76 & 48.59 & 73.46 & 71.93 & 72.68 \\
%    \bottomrule
%    \end{tabular}} 
%     \label{tab:mainhard}
%     \caption{Extraction Result on complicated extraction case with different extraction order decision. HacRED is tested on at least 5 triples of the same relation.}
%\end{table*}

\begin{table*}[!htb]
    \begin{minipage}[!htb]{0.48\textwidth}
    \makeatletter\def\@captype{table}\makeatother
    \centering
    \resizebox{\columnwidth}{!}{\begin{tabular}{lrrrrrr}
    \toprule
         & \multicolumn{3}{c}{SKE21} & \multicolumn{3}{c}{HacRED} \\
         \cmidrule{2-7}       & \multicolumn{1}{c}{Prec.} & \multicolumn{1}{c}{Reca.} & \multicolumn{1}{c}{F1} & \multicolumn{1}{c}{Prec.} & \multicolumn{1}{c}{Reca.} & \multicolumn{1}{c}{F1} \\
        
    \midrule
     Sequence order & \textbf{94.74} & 81.08 & 87.38 & \textbf{66.62} & 78.67 & 72.15 \\
     Random order & 92.75 & 80.63 & 86.27 & 65.94 & 79.41 & 72.05 \\
     Adaptive order & 93.37 & \textbf{82.43} & \textbf{87.56} & 66.31 & \textbf{80.56} & \textbf{72.74} \\
    \bottomrule
    \end{tabular}}
     \caption{Extraction Result on complicated extraction case with different extraction order decision. HacRED and SKE21 are both tested on at least 5 triples of the same relation.}
    \label{tab:mainhard1}
    \end{minipage}
    \hspace{5mm}
    \begin{minipage}[!htb]{0.48\textwidth}
    \makeatletter\def\@captype{table}\makeatother
    \centering
    \resizebox{\columnwidth}{!}{\begin{tabular}{lrrrrrr}
    \toprule
         & \multicolumn{3}{c}{DuIE} & \multicolumn{3}{c}{DuEE} \\
         \cmidrule{2-7}       & \multicolumn{1}{c}{Prec.} & \multicolumn{1}{c}{Reca.} & \multicolumn{1}{c}{F1} & \multicolumn{1}{c}{Prec.} & \multicolumn{1}{c}{Reca.} & \multicolumn{1}{c}{F1} \\
        
    \midrule
     Sequence order & 44.88 & 48.67 & 46.70 & 72.41 & 69.30 & 70.82 \\
     Random order & 44.25 & 47.91 & 46.01 & 72.73 & 70.89 & 71.80 \\
     Adaptive order & \textbf{46.60} & \textbf{50.76} & \textbf{48.59} & \textbf{73.88} & \textbf{71.69} & \textbf{72.76} \\
    \bottomrule
    \end{tabular}}
     \caption{Extraction Result on complicated extraction case with different extraction order decision. DuIE is restricted in at least 3 roles and DuEE is restricted in at least 5 roles.}
    \label{tab:mainhard2}
    \end{minipage}
\end{table*}

\section{Experiments}
\subsection{Datasets}
We evaluate our methods on several public and accessible complicated information extraction datasets, including NYT, NYT10-HRL, SKE21, HacRED, DuIE, DuEE, which are challenging for many novel extraction methods. We give brief introduction to these dataset in appendix.

\subsection{Comparing methods and Metrics}
% Methods
We compare our methods with several models on the same dataset, including NovelTagging\cite{zheng2017joint}, CoType\cite{ren2017cotype}, HRL\cite{takanobu2019hierarchical}, and recent PLM-based model
 CasRel\cite{wei2020novel}, TPlinker\cite{wang2020tplinker} and \rere\cite{xie2021revisiting}. 

% Metric
We choose the exact match that an extracted relational triple \textit{(subject, relation, object)} is regarded as correct only if the relation and the spans of both subject and object are correct. We report the standard micro precision (Prec.), recall (Reca.) and F1 scores for the relation extraction experiments. And for the event extraction task (only DuEE in our experiment), we report the word-level metrics, which considers the correctness of the arguments in word level. We give the detail of this metric in appendix.

\begin{table*}[tb]
    \centering
    \begin{tabular}{p{3.3cm}p{5.9cm}p{5.5cm}}
        \toprule
        \textbf{Instance} & \textbf{Extraction Process} & \textbf{Extraction Result}\\
        \midrule
         \makecell[l]{\underline{\textbf{Instance \#1}}  \\ On October 14, Redmi \\officially released two \\new entry-level smart-\\phones, Redmi 8 \\and Redmi 8A.} & 
         \begin{minipage}[b]{6cm}
		  \centering
		  \raisebox{-.5\height}{\includegraphics[width=\linewidth]{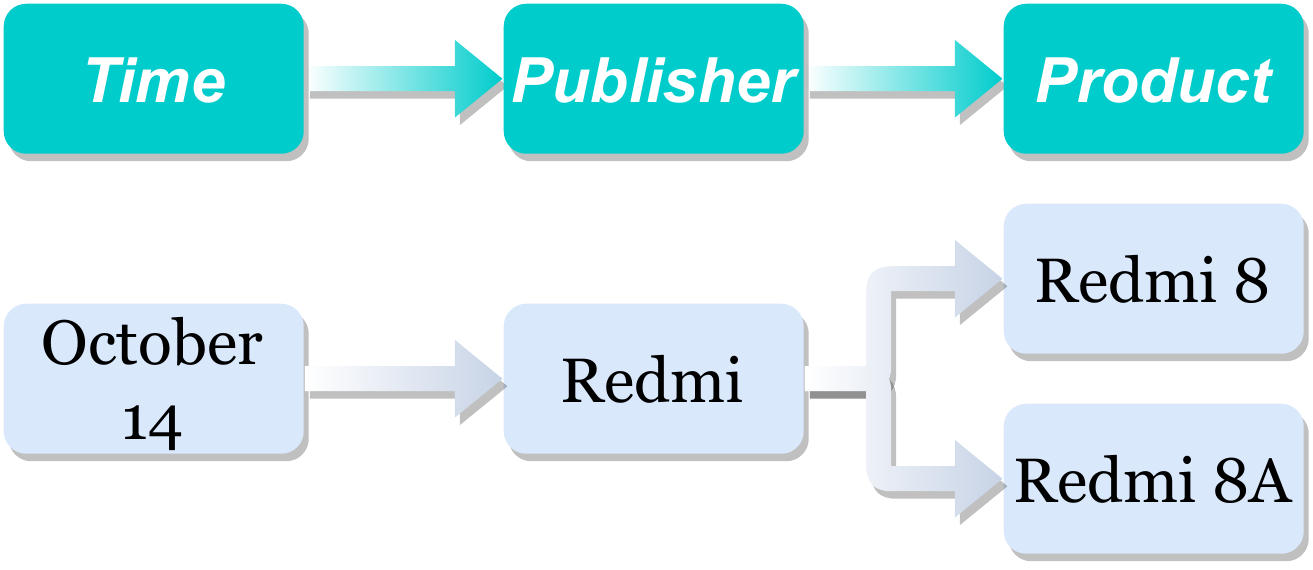}}
	\end{minipage} & 
        \makecell[l]{\textit{\textbf{Time}}: October, \\ \textit{\textbf{Publisher}}: Redmi, \\\textit{\textbf{Product}}: Redmi 8A; \\ \textit{\textbf{Time}}: October, \\ \textit{\textbf{Publisher}}: Redmi, \\\textit{\textbf{Product}}: Redmi 8A} \\
         \midrule
         \makecell[l]{\underline{\textbf{Instance \#2}} \\Huami Technology re-\\leases Midong Health \\Watch and AMAZFIT \\Smart Watch 2.} & 
         \parbox[c]{6cm}{\includegraphics[width=\linewidth]{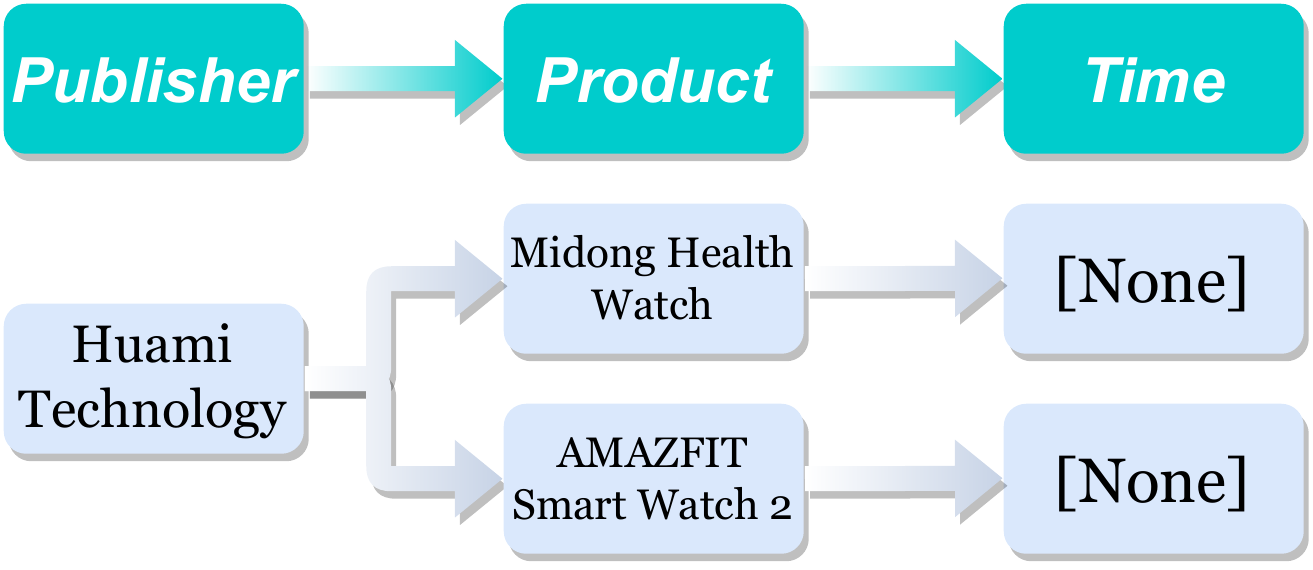}}
          & \makecell[l]{\textit{\textbf{Time}}: [None], \\\textit{\textbf{Publisher}}: Huami Technology, \\\textit{\textbf{Product}}: Midong Health Watch; \\\textit{\textbf{Time}}: [None], \\\textit{\textbf{Publisher}}: Midong Health Watch, \\\textit{\textbf{Product}}: AMAZFIT Smart Watch 2}\\
         \bottomrule
    \end{tabular}
    \caption{Instance of extracting complicated schema through dynamically assigning extraction order.}
    \label{tab:case}
\end{table*} 

\subsection{Main Results}
Because we only consider the extraction order assignment in every instance, we add a classification module to first recognize the relation in sentence, and then extract the subject and object with the extraction order RL agents assign. According to the result in Table \ref{tab:main}, compared to the main stream relation extraction methods, ours method achieves improvement on both precision, recall and F1 score.

\subsection{Extraction order}
To demonstrate the effectiveness of our methods, we also conduct an experiment on different extraction order decision strategy on more challenging event argument extraction task. In this experiment, we mainly focus on the performance of extraction result in different extraction order, so ground-truth schema (relation) for every instance would be offered beforehand, we only test the correctness on the roles. We also provide the result of extracting arguments in a pre-defined order and random order, as baselines. Table \ref{tab:order} shows the result of different extraction order in different dataset, and our method achieve the best in every dataset. Compare to the standard relation extraction task, our method perform better on complex information extraction task (DuIE and DuEE).

\subsection{Complicated Extraction settings}
To further demonstrate the advantage of dynamically order extraction decision by RL, we conduct experiment with more complex extraction tasks, which contains more tuples or more arguments. For the former, we limit the minimum number of extraction tuples, and we limit the minimum number of extraction roles for the latter. 

% 卡长度 or 抽取槽位个数来展示面对进一步复杂的抽取任务效果更佳
Table \ref{tab:mainhard1} and \ref{tab:mainhard2} show that compared to fixed order extraction or random order extraction method, our framework has a more significant improvement over the original metrics. This is intuitive and reasonable, the extractor is more sensitive to the extraction order in more complex sentence. Besides, compare to the Table \ref{tab:mainhard1} and \ref{tab:mainhard2}, we can find that our method improves the latter settings more significantly. It is because increasing the number of tuples does not increase the length of the extracted path, but only increase the difficulty of single-step extraction by the extractor. In contrast, the increase of the role number leads to an increase in the length of the extraction path, which makes the decision of extraction order more difficult. The results once again prove that the extraction order matters in complicated extraction.

% 相较于元组的个数，槽位更多越依赖抽取的顺序，这是因为推理路径的长度问题

\subsection{Case Study}
% 用RL预测的结果说明模型预测的可解释性，并证明模型针对不同的句子，具有不同的抽取顺序的选择，是真正实现Dynamic Extraction的能力。
With taking RL agent into consideration, we can easily observe the extraction order in different instance. Table \ref{tab:case} show the instances that the extraction process in different sentence. Though two instances share the same event schema \textit{Product release}, the RL agent assign different extraction order dynamically. The first sentence contains an obvious element of time, while the second does not, so our methods put the extraction order of time from the first to the last. The case strongly demonstrates the effectiveness of our method.

\section{Conclusion}
In this paper, we propose a novel adaptive order IE paradigm to find the optimal element extraction order for different instances. We propose a RL-based framework to generate optimal extraction for each instance dynamically and a co-training framework adapted to RL to alleviate the exposure from the extrator. Extensive experiments show that our proposed method can beat previous methods and effectively improve the performance of various IE tasks, especially for complex ones.

\section*{Acknowledgements}
This research is funded by National Key Research and Development Project (No. 2020AAA0109302), National Natural Science Foundation of China (No. 62102095, 62072323), Shanghai Science and Technology Innovation Action Plan (No. 22511104700, 22511105902), Shanghai Municipal Science and Technology Major Project (No.2021SHZDZX0103), and Science and Technology Commission of Shanghai Municipality Grant (No. 22511105902).

\section*{Limitations}
% Inference speed
Despite the remarkable improvement on complicated information extraction, there are still some limits of our method. 

First, due to the multi-round argument extraction modeling, we discard the parallelism in element extraction. Furthermore, the MDP process interacting with the DQN further increases the computational load of the extraction process. So compare to other methods, our framework is relative slow at the inference stage.
% Without Predicate

Second, though our framework can be easily adapted to different extraction task with different schema, we still need an extra module helping identifying the relations (event types) in the instance beforehand. Because of the difference task definition and modeling (extraction task and classification task), although recognizing them potentially implies order decision making, they are beyond the scope of this paper. 

\section*{Ethics statement}
We hereby declare that all authors of this article are aware of and adhere to the provided ACL Code of Ethics and honor the code of conduct.
\paragraph*{Use of Human Annotations}
Human annotations are only utilized in the early stages of methodological research to assess the feasibility of the proposed solution. All annotators have provided consent for the use of their data for research purposes. We guarantee the security of all annotators throughout the annotation process, and they are justly remunerated according to local standards. Human annotations are not employed during the evaluation of our method.

\paragraph*{Risks}

The datasets used in this paper have been obtained from public sources and anonymized to protect against any offensive information. Though we have taken measures to do so, we cannot guarantee that the datasets do not contain any socially harmful or toxic language.

\newpage
\bibliography{anthology,custom}
\bibliographystyle{acl_natbib}

\newpage
\appendix

\section{Dataset Introduction}
\label{sec:appendix}
\paragraph{NYT} NYT~\cite{riedel2010modeling} is the very early version of NYT series dataset. It is based on the articles in New York Times, contains $66,194$ text and 24 types of relation.

\paragraph{NYT10-HRL} NYT10-HRL is an improved version of the origin NYT dataset. After preprocessed by HRL~\cite{takanobu2019hierarchical} that removed training relation not occurring in the testing test and "NA" sentence, NYT10-HRL contains $70,339$ sentences for training and $4,006$ sentences for test.

\paragraph{DuIE} DuIE\cite{li2019duie} is a Chinese information extraction dataset. It contains more than $210,000$ sentences and $48$ pre-defined schema gathered from Baidu Encyclopedia, Baidu Tieba and Baidu Information Stream. Compared to the previous version, it contains $5$ Multiple-O-values schema (Schema with multiple object slots and values), which greatly increase the difficulty of the task.

\paragraph{HacRED} HacRED~\cite{cheng2021hacred}\footnote{\url{https://github.com/qiaojiim/hacred}} is a novel challenging extraction dataset, which consists of $65,225$ relational facts annotated from $9,231$ wiki documents with sufficient and diverse hard cases

\paragraph{DuEE} DuEE~\cite{li2020duee} is a Chinese document-level event extraction dataset. It contains $11,224$ documents categorized into 65 event types, along with $41,520$ event arguments mapped to $121$ argument roles, which is the largest Chinese EE dataset. 

\paragraph{SKE21} SKE2019 is the largest Chinese dataset available for relation extraction publish by Baidu, which contains $194,747$ sentences for training. \cite{xie2021revisiting} manually labeled $1,150$ sentences from test set with $2,765$ annotated triples.

\section{Experiment details}
Our experiments are conducted on single RTX3090 GPUs. All deep models, including the extraction model and decision model, are implemented using the PyTorch framework. We initialized the model with the bert-base-cased and chinese-roberta-wwm-ext respectively, training 10 epochs for both the extractor and classifier. As for the reinforcement learning module, we set the buffer size to 100,000 and target network update step at 200. We trained 5 epochs on the standard relation extraction task and 10 epochs for DuIE and DuEE. Additionally, the exploration parameter $\epsilon$ was initialized at 0.9 and the discount factor $\gamma$ was set to 0.5 within the DQN framework. To encourage continuous exploration, the exploration rate became 0.9 times of its own in a certain steps until it reached 0.05. We calculated the necessary updating steps using the following formula.
\begin{equation}
    \begin{aligned}
    \# Total\_Steps &= \#Epoch \times \#Datasets \\
    \# Update\_steps &= \lfloor \frac{\#Total\_Steps * \log{0.95}}{\log{0.05} - \log{0.9}} \rfloor
    \end{aligned}
\end{equation}
We employed AdamW optimizer as the optimizer, with using linear scheduling with warming up proportion 10\%. More details are listed in Table \ref{tab:hyperpara}.

\begin{table}[tb]
    \centering
    \resizebox{\columnwidth}{!}{\begin{tabular}{ccc}
        \toprule
        \textbf{Hyper-parameters} & Extractor & Decision Model\\
        \midrule
        Batch size & 48 & 32 \\
        Learning rate & 1e-5 & 1e-5\\
        Max length & 512 & 512 \\
        \bottomrule
    \end{tabular}}
    \caption{Hyper-parameters for training extractor and agent model respectively.}
    \label{tab:hyperpara}
\end{table} 

\section{Word-level Metric}
We offer the word-level precision, recall and F1 score calculation formula. First we calculate the F1 score of a single arguments.
\begin{gather}
    Set_a = Set_p \cap Set_g, \\
    Text P = \frac{\left| Set_a \right|}{\left| Set_p \right|},
    Text R = \frac{\left| Set_a \right|}{\left| Set_g \right|} \\
    Text F1 = \frac{2\cdot Text P \cdot Text R}{Text P + Text R}
\end{gather}
where $Set_p$ denotes the set of words in predict argument, and $Set_g$ denotes the set of words in ground truth argument. We set $Text F1 = 1$ if both $Set_a$ and $Set_b$ are empty set. 

For every predict event $E_p$ and ground truth event $E_g$, we calculate their match score through calculate their mean $Text F1$ score.
\begin{equation}
    Score(E_p, E_g) = \frac{1}{N}\sum_N TextF1(arg_p, arg_g)
\end{equation}
where $N$ denotes the number of the roles that event schema contains.

We separately calculate the precision for prediction event and recall for ground truth event.
\begin{equation}
    \begin{aligned}
        Prec_E(E_p) & = \max_{e \in \textbf{E}_g} Score(E_p, e)\\
        Reca_E(E_g) & = \max_{e \in \textbf{E}_g} Score(e, E_g)
    \end{aligned}
\end{equation}
where $\textbf{E}_p$ and $\textbf{E}_g$ are the predict and golden event set of the same instance. 

Finally, we regard the mean precision and recall of every prediction and golden annotations in the whole test set as our Prec. and Reca.
\begin{equation}
    \begin{aligned}
        Prec &= \frac{1}{N_p} \sum_{E_p} Prec(E_p)\\
        Reca &= \frac{1}{N_g} \sum_{E_g} Reca(E_g)\\
        F1 &= \frac{2\cdot Prec \cdot Reca}{Prec + Reca}
    \end{aligned}
\end{equation} 
\end{document}